# OBJECT-ORIENTED DYNAMIC NETWORKS

## Dmytro Terletskyi, Alexandr Provotar

*Abstract*: *This paper contains description of such knowledge representation model as Object-Oriented Dynamic Network (OODN), which gives us an opportunity to represent knowledge, which can be modified in time, to build new relations between objects and classes of objects and to represent results of their modifications. The model is based on representation of objects via their properties and methods. It gives us a possibility to classify the objects and, in a sense, to build hierarchy of their types. Furthermore, it enables to represent relation of modification between concepts, to build new classes of objects based on existing classes and to create sets and multisets of concepts. OODN can be represented as a connected and directed graph, where nodes are concepts and edges are relations between them. Using such model of knowledge representation, we can consider modifications of knowledge and movement through the graph of network as a process of logical reasoning or finding the right solutions or creativity, etc. The proposed approach gives us an opportunity to model some aspects of human knowledge system and main mechanisms of human thought, in particular getting a new experience and knowledge.*

*ACM Classification Keywords*: *E.2 Data Storage Representations — Object representation, D.3.3 Language Constructs and Features — Abstract data types, Classes and objects, Data types and structures, D.1.5 Object-oriented Programming, F.4.1 Mathematical Logic — Set theory.*

## Introduction

Modern AI includes many directions, which differ from each other, but at the same time they have something in common. One of the main targets for researchers of all these directions is the development of intelligent information systems (IIS) for solving particular practical problems in corresponding areas. Nowadays, development of IIS is often reduced to heuristic programming. Nevertheless, there is a variety of IIS, which are based on knowledge representation model (KRM). The most famous and common are Semantic Nets, Conceptual Dependency, Frames, Scripts, Logical and Production Models, Ontologies, etc. Each of these models has own specifics and is useful in the particular domain. However, we need to implement a KRM while developing certain IIS. That is why our IIS is going to have at least two levels: the level of KR model and the level of its practical implementation. Sometimes, implementation of certain KRM can cause additional problems and difficulties, which are connected with interaction between two abstraction levels of IIS. This situation led to the development of logical programing and such programming language as Prolog, where KRM is integrated within the language. Such approach gives us an opportunity to represent knowledge using just programming language, because corresponding KRM is integrated within it.

Clearly, the development of software for solving particular tasks, management by some process, etc. is easier than the development of software, which has some level of individuality and intellectuality and can do more complicated tasks than just some computations. That is why questions about useful and powerful tool for such development appear. Modern programming includes many different paradigms, approaches, techniques and programming languages. Object-Oriented Programming (OOP) is one of the famous, useful and powerful programming paradigms nowadays. Indeed, according to [Langpop; Tiobe; Sourseforge] the most popular programming languages in 2013 were languages, which support OOP. Furthermore, many of programming languages have been extended to object-oriented languages: C to C++; Prolog to Object Prolog; COBOL to Object COBOL; SQL to OQL; and LISP to COOL [Sowa, 2000]. That is why questions about usefulness of OOP approach for development IIS, which are based on some KRM, appear.

## Object-Oriented Knowledge Representation Models

As it was mentioned before, there are many different types of KR models. However, as in the case with Prolog, if we want to use an OOP language, as a tool for implementation of IIS, we need to use certain object-oriented KR models as a base for them. Nowadays, the most known and developed object-oriented KR models are *frames* and *scripts*, that is why let us consider some their basic and important aspects.

**Frames.** Generally, frame consists of set of slots, where the corresponding value is assigned for each slot. Every slot has some filler for itself and in such a way provides information about one of the frame's attributes. Furthermore, some fillers can be frames. According to [Brachman, Levesque, 2004], there are two types of frames: *individual frames,* which are used for representation of single objects, and *generic frames,* which are used for representation of classes of objects. There are special slots as *instance-of*, *is-a*, *a-kind-of*, etc., which help to organize the relations between different frames and types of frames, in particular between individual frames and generic frames, and in such a way to build up the frame system.

In addition, frames have some methods associated with them, which are called *procedures* or *procedural attachments*. Each procedure is a set of some instructions, associated with a frame that can be executed on request. Particular examples of procedural attachments are *slot-reader*, *slot-writer*, etc. Other important procedures are instance constructors, which create instances of classes. Such procedures called when-needed or if-needed procedures and can execute only when they are really required.

Frames have a powerful tool for creating new knowledge called inheritance. It means that frames can inherit the attributes of other frames in the hierarchical structure. Such kinds of slots as instance-of and is-a play an important part in this process, in particular they fill individual frames using other individual or generic frames. Furthermore, frames have such tool as multiple inheritance, which give an opportunity for the frame to inherit properties from more than one other frame.

Analyzing structure of the frames, we can conclude about such their advantages as ability to be represented in the form of a table; to store and use default values in the reasoning process; ability to be structured hierarchically and thus allow easy classification of knowledge; to combine procedural and declarative knowledge using one knowledge representation scheme; to constrain allowed values, or make values be entered within a specific range [Kendal, Creen, 2007]. In addition, frames make semantic nets more powerful by allowing complex objects to be represented as a single frame, rather than as a large network structure. It also provides a common way to represent stereotypic entities, classes, inheritance, and default values [Luger, 2008]. Frame systems tend to have a centralized, conventional control regime, whereas OOP systems have objects acting as small, independent agents sending each other messages, that is why, there can be some applications for which a frame-based system can provide some advantages over a more generic OOP system [Brachman, Levesque, 2004]. Likewise, frames can be used as a data structure for Expert Systems [Coppin, 2004].

However, frames also have some disadvantages. They do not provide the most efficient method to store data for a computer; can lead to "procedural fever"; require care in the design stage to ensure that suitable taxonomies [Kendal, Creen, 2007]. In addition, frames can be considered as a simplified version of a semantic network where only is-a relationships are applied. That is why, the inheritance of default properties of frames in a hierarchy leads to problems with *exceptions* and *multiple inheritance*. First problem arises when a property of a supertype applies to most but not all of its subtypes. The second one arises when a particular subtype may have more than one supertype from which it can inherit properties, which may conflict [Way, 1991; Coppin, 2004].

**Scripts.** The scripts are structured representation describing stereotyped sequences of events in a particular context. They are frame-like structures for organizing conceptual dependency structures into descriptions of typical situations. Scripts consist of *entry conditions* that must be true for the script to be called; *results* that are true once the script has terminated; *props* that support the content of the script; *roles* that the individual participants perform; *scenes* that presents a temporal aspect of the script.

Frames and scripts are particularly appealing as means for knowledge representation because psychological studies have shown that people tend to rely on knowledge from previous experience whenever possible, and they use this knowledge and adapt it to handle new or slightly different situations. Therefore, instead of analyzing and building descriptions of each new situation as it occurs, people draw on a large collection of structures, which represents their previous experience with objects, people, and situations, and use these past expectations to guide them in analyzing and representing new experiences [Way, 1991].

Nevertheless, scripts, like other KRM have certain problems, in particular the *script match* problem and *between-the-lines* problem. The first problem is that not exists algorithm, which can guarantee correct choices of script in particular situation. The second problem is that not possible to know beforehand the possible occurrences that can break a script. These problems are not unique to script technology but are inherent in the problem of modelling semantic meaning [Graham, Barrett, 1997].

Despite all advantages and disadvantages of frames, which were mentioned above, we can conclude, that frames help us to describe and somehow to represent relations between objects and classes, which are represented through relations with other objects and classes, using inheritance and special slots as *instance-of*, *is-a*, *a-kind-of*, etc. However, such representation of objects and classes does not describe their properties and types without additional information as links with other objects and classes. It means, for representation of some objects we need to represent a lot of other objects and classes, which have higher level in the particular hierarchy of frames. Thus, logical reasoning within particular structure of frames reduces to manipulations with its hierarchy. Nevertheless, questions about what is a starting point in the logical reasoning, how many frames we need for such reasoning, what is superclass in the hierarchy, etc. appear.

Concerning scripts, we can conclude that they have similar to frames nature, but at the same time, they are used for representation of sequences of actions in particular locations. The main feature of scripts is a representation of possible scenarios in the certain locations. However, they do not pay enough attention to features of certain location, in particular objects, which form it. Nevertheless, question about how many related scripts we need for managing by different situations in certain location also appears.

## Objects and Classes

We can represent different objects and classes, using frames and scripts. The difference between them is that they represent different types of objects and classes. Frames represent relations between some objects and classes, creating a hierarchy in such a way. Scripts represent actions and relations between them, creating some scenarios. However, both of them do not describe features of their objects and classes and thus do not express fully their semantics. In contrast to frames and scripts, OOP pays more attention to description of features of objects and classes, herewith also creating some hierarchy. That is why, let us consider some features of OOP and try to figure out what an object and a class is.

In OOP object and class are the main concepts. Objects are the building blocks for object-oriented programs. We associate these blocks with the objects of real world, during developing programs. Every object is defined by two terms: attributes and behaviors. Attributes are properties of object, which describe it, and behaviors are procedures, functions (methods) which we can apply to this object and change its state, form and so on [Weisfeld, 2008]. Real world consists of objects, and OOP is the approach for description and simulation of this world or some its particular parts [Pecinovský, 2013].

Let consider such object as "natural number". It is clear that every natural number must be integer and positive. These are characteristic properties of natural numbers. It is obvious, that 12 is really a natural number, but −1 and 7.32, for example, are not natural numbers. While analyzing this fact, we can conclude that each object has certain properties, which define it as some essence. In contrast to OOP, generally properties of objects can be divided into two types – quantitative and qualitative. We are going to define these two types of object's properties formally, but their semantics has intuitive nature.

**Definition 1.** *Quantitative property of object* $A$ *is a tuple* $p_i(A) = (v(p_i(A)), u(p_i(A)))$ *where* $i = \overline{1,n}$, $v(p_i(A))$ *is an quantitative value of* $p_i(A)$ *and* $u(p_i(A))$ *are units of measure of quantitative value of* $p_i(A)$.

**Example 1.** Suppose we have a car and one of its properties is speed. We can present this property as $p_s(Car) = (v(p_s(Car)), u(p_s(Car)))$ and if our car has speed 150 km/hour, then property is the following $p_s(Car) = (150, km/hour)$.

**Definition 2.** *Two quantitative properties* $p_i(A)$ *and* $p_j(B)$ *where* $i = \overline{1,n}$, $j = \overline{1,m}$ *are equivalent, i.e.* $Eq(p_i(A), p_j(B)) = 1$, *if and only if* $u(p_i(A)) = u(p_j(B))$.

**Definition 3.** *Qualitative property of object* $A$ *is a verification function* $p_i(A) = vf_i(A)$, $i = \overline{1,n}$ *which is defined as a mapping* $vf_i(A) : p_i(A) \rightarrow [0,1]$.

**Example 2.** Suppose we have a natural number $n$, and one of its properties is positivity. We can present this property as follows $p_{pos}(n) = vf_{pos}(n)$, where $vf_{pos}(n)$ is verification function of property $p_{pos}(n)$. In this case, function *is defined as a mapping* $vf_{pos}(n) : p_{pos}(n) \rightarrow \{0,1\}$, and it is a particular case of verification function – predicate or *Boolean-valued function.*

**Definition 4.** *Two qualitative properties* $p_i(A)$ *and* $p_j(B)$ *where* $i = \overline{1,n}$, $j = \overline{1,m}$ *are equivalent, i.e.* $Eq(p_i(A), p_j(B)) = 1$, *if and only if* $(vf_i(A) = vf_j(A)) \wedge (vf_i(B) = vf_j(B))$.

**Definition 5.** *Specification of object* $A$ *is a vector* $P(A) = (p_1(A), ..., p_n(A))$ *where* $p_i(A)$, $i = \overline{1,n}$ *is quantitative or qualitative property of object* $A$.

**Definition 6.** *Object is a pair* $A / P(A)$, *where* $A$ *is object's identifier and* $P(A)$ *is a specification of object.*

Essentially, object is a carrier of some properties, which define it as some essence, and help us recognize it among other objects.

**Definition 7.** *Two objects* $A$ *and* $B$ *are similar, if and only if they have the same properties and behavior, i.e.* $P(A) = P(B)$ *and* $F(A) = F(B)$.

Besides properties of objects, we should allocate operations (methods) which we can apply to objects, considering the features of their specifications. That is why, it will be useful to define concept of object's operation (method).

**Definition 8.** *Operation (method) of object $A$ is a function $f(A)$, which we can apply to object $A$ considering the features of its specification.*

Definition of method of object is similar to corresponding definition in OOP. However, there is a difference between them. Usually methods in OOP are functions, which we can execute for objects. In contrast to OOP, we divide methods of objects on two types, depending on character of their action: *modifiers* and *exploiters*. Modifiers are functions, which can change objects, in particular some fields of objects. Exploiters are functions, which use objects as arguments and cannot change them.

**Example 3.** Let us consider such objects as natural numbers $n$, $m$. Operations of sum and multiplication, i.e. $f_1(n,m) = n + m$ and $f_2(n,m) = nm$ are the simplest examples of exploiters for them.

**Example 4.** Let us consider integer number $k$. Incrementation operation $f_1(k) = k + w$ is the simplest example of modifier.

**Example 5.** Other simple examples of modifiers and exploiters are $get(Object)$ and $set(Object, Value)$ functions, which are common in many OO languages.

**Definition 9.** *Signature of object $A$ is a vector $F(A) = (f_1(A), ..., f_m(A))$, where $f_i(A)$, $i = \overline{1, m}$ is an operation (method) of object $A$.*

Generally, we can divide objects on concrete and abstract, and does not matter when or how someone has created each particular object. It is implementation of its abstract image – a prototype, which is essentially an abstract specification for creating the future particular objects. In other words, classes are blueprints, which we use as the basis for objects building [Weisfeld, 2008]. In OOP class consists of fields and methods. Fields form specification of class and methods are functions, which we can apply to objects of this class. We will define concept of class of objects using corresponding idea of OO class.

**Definition 10.** *Class of objects $T$ is a tuple $T = (P(T), F(T))$, where $P(T)$ is abstract specification of some quantity of objects, and $F(T)$ is their signature.*

When we talk about class of objects, we mean properties of these objects and methods, which we can apply to them. In other words, class of objects is a generalized form of consideration of objects and operations on them, without these objects. In OOP every particular object has the same fields and behavior as its class, i.e. it has the same specification and signature. According to this, we can define concept of homogeneous class of objects.

**Definition 13.** *Homogeneous class of objects $T$ is a class of objects, which contains only similar objects.*

**Example 6.** The simplest examples of homogeneous classes of objects are class of natural numbers, class of letters of English alphabet, class of colors of the rainbow, etc.

There are many different objects of real world, which belong to different classes, and if we need to work with them, we can describe them, using new separate homogeneous class for each new type of objects. Especially, if we work with not very big quantity of different types of objects, we can do it without any fears. However, if we need to work, for example, with a few thousands of different types or more, just a process of description of such types is very complex and time-consuming not to mention size of code and performance of such programs. Nevertheless, besides homogeneous classes there are inhomogeneous classes of objects, which describe objects of different types within one class. It means that each object of such class can have different properties and methods.

**Definition 14.** *Inhomogeneous class of objects* $T$ *is a tuple* $T = (Core(T), pr_1(A_1), ..., pr_n(A_n))$ *, where* $Core(T) = (P(T), F(T))$ *is the core of class of objects* $T$ *, which includes only properties and methods similar to corresponding properties of specifications* $P(A_1), ..., P(A_n)$ *and corresponding methods of signatures* $F(A_1), ..., F(A_n)$ *respectively, and where* $pr_i(A_i) = (P(A_i), F(A_i))$ *,* $i = \overline{1, n}$ *are projections of objects* $A_1, ..., A_n$ *, which consist of properties and methods typical only for these objects.*

**Example 7.** Let us consider such class of objects as natural numbers $N$. Clearly that it is a member of such classes as integer numbers, rational numbers and real numbers simultaneously, i.e. $N \in Z \in Q \in R$. As we can see, class $R$ is the biggest in this case. Furthermore, it consists of objects of different types that contradicts concept of OO class. Of course, in programing languages such types are basic and are built in language. However, in OOP we need to use separate class for description of each such class, because different objects from one OO class cannot have different specifications and signatures. That is why we cannot describe such classes of objects using only one OO class.

## Operations on Objects and Classes of Objects

As it was mentioned above, in OOP objects have methods, however, majority of them are local with respect to objects, and cannot be applied to objects of different types, i.e. they are not polymorphic. Of course, there are some methods, which we can apply to objects of different types, but usually we need to use overloading operators for it [Stroustrup, 2013]. Nevertheless, union, intersection, difference, symmetric difference and cloning operations on objects and classes of objects, were proposed in [Terletskyi, Provotar, 2014]. These operations have set-theoretic nature and they are universal in this sense as they can be applied to any objects and classes of objects regardless of their features. We will not concentrate much attention on these operations and just will consider some examples of their using.

**Example 8.** Let us consider such geometrical objects as triangle, square and trapeze. It is obvious, that these objects belong to different classes of convex polygons. Let us denote triangle as $A$ square as $B$, trapeze as $C$ and describe their classes as follows

$$T(A) = (P(A), F(A)) = ((p_1(A), ..., p_5(A)), (f_1(A), f_2(A))),$$
$$T(B) = (P(B), F(B)) = ((p_1(B), ..., p_5(B)), (f_1(B), f_2(B))),$$
$$T(C) = (P(C), F(C)) = ((p_1(C), ..., p_5(C)), (f_1(C), f_2(C))).$$

The meaning of each property and method is defined by the Table 1.

*Table 1. Meaning of properties and methods of figures* $A, B, C$

| Properties/Methods | Meaning |
|---|---|
| $p_1(A), p_1(B), p_1(C)$ | quantities of sides of figures |
| $p_2(A), p_2(B), p_2(C)$ | sizes of sides of figures |
| $p_3(A), p_3(B), p_3(C)$ | quantities of angles of figures |
| $p_4(A), p_4(B), p_4(C)$ | measure of angles of figures |
| $p_5(A)$ | triangle inequality |
| $p_5(B)$ | parallelism of opposite sides of figure |
| $p_5(C)$ | parallelism of two sides of figure |
| $f_1(A), f_1(B), f_1(C)$ | functions of perimeter calculation of figures |
| $f_2(A), f_2(B), f_2(C)$ | functions of area calculation of figures |

Specifications and signatures of objects $A, B, C$ can include more or less properties and methods, than we have presented in Table 1, everything depends on the level of detail.

**Union.** Let us apply the union operation to objects $A, B, C$ and create new set of objects.

$$S = A / T(A) \cup B / T(B) \cup C / T(C) = \{A, B, C\} / T(S)$$

We have obtained a new set of objects $S$ and a new class of objects

$$T(S) = (Core(T(S)), pr_1(A), pr_2(B), pr_3(C)),$$

where $Core(T(S)) = (p_1(T(S)), p_2(T(S)), p_3(T(S)), p_4(T(S)), f_1(T(S)))$, property $p_1(T(S))$ is a quantity of sides of figures, property $p_2(T(S))$ is size of sides of figures, $p_3(T(S))$ is a quantity of angles of figures, $p_4(T(S))$ is measure of angles of figures, method $f_1(T(S))$ is a function of figures' perimeter calculation and $pr_1(A) = (p_5(A), f_2(A))$, $pr_2(B) = (p_5(B), f_2(B))$, $pr_3(C) = (p_5(C), f_2(C))$. Essentially, a set of objects $S$ is the set of triangles of class $A$, squares of class $B$ and trapezes of class $C$. Class of objects $T(S)$ describes three types of geometrical figures $T(A)$, $T(B)$ and $T(C)$.

**Intersection.** Let us calculate intersection of triangle $A$ and square $B$.

$$A / T(A) \cap B / T(B) = T(A \cap B)$$

As the result, we have obtained new class of objects $T(A \cap B)$, which does not contain any projections of objects, i.e. $T(A \cap B) = (Core(T(A \cap B)))$, where

$$Core(T(A \cap B)) = (p_1(T(A \cap B)), p_2(T(A \cap B)), p_3(T(A \cap B)), p_4(T(A \cap B)), f_1(T(A \cap B))).$$

Meaning of all properties and methods of $Core(T(A \cap B))$ is definitely the same as in the case of union. Class of objects $T(A \cap B)$ describes some type of geometrical figures. However, we do not know exactly which one, but it consists of properties and methods which are simultaneously common for triangle $A$ and square $B$. Moreover, this class of objects is a part of description of any convex polygon, because each polygon has sides and angles.

**Difference.** Let us calculate difference of triangle $A$ and trapeze $C$.

$$A / T(A) \setminus C / T(C) = T(A \setminus C)$$

As the result, we have obtained new class of objects $T(A \setminus C)$, which does not contain core, i.e. $T(A \setminus C) = (pr_1(A))$, where $pr_1(A) = (p_5(A), f_2(A))$. The obtained new class $T(A \setminus C)$, describes, unlike the previous case, the concrete geometric figure - triangle, but using less specification, than given in the Table 1. **Symmetric Difference.** Let us calculate symmetrical difference using the same figures, i.e. triangle $A$ and trapeze $C$.

$$A / T(A) \div C / T(C) = T(A \div C)$$

As the result, we have obtained a new class of objects $T(A \div C)$, which as in the previous case, does not contain core, i.e. $T(A \div C) = (pr_1(A), pr_2(C))$, where $pr_1(A) = (p_5(A), f_2(A))$ and $pr_2(C) = (p_5(C), f_2(C))$. The class of objects $T(A \div C)$, describes two types of geometrical figures, one of them is a triangle, another one is ambiguously defined.

**Cloning.** Let us apply cloning operation to triangle $A$.

$$Clone_1(A) = A_1 / T(A)$$

As the result, we have obtained a new indexed copy of triangle $A$.

Clearly, that these five operations are exploiters, by using which we can create new classes of objects and sets of objects. It is directly connected to creation of the inhomogeneous classes of objects that means obtaining of new knowledge, in particular classes of these sets. The Example 8 illustrates the basic ideas of these operations in a context of their application to classes of objects. In addition, they can be extended for objects [Terletskyi, 2013].

Creation of new classes of objects, using proposed operations, is directly connected to process of Runtime Class Generation (RCG) in OOP. The main idea of RCG is an opportunity to obtain new classes of objects during program execution. Nowadays, there are few approaches for implementation of this task for some OOP languages, in particular [CGLib] for Java and [CodeDOM] for C#. However, these tools are implemented for such platforms of programming as Java and .NET and based on manipulating with bytecode, which limits their application.

One of the main tools for class creation in OOP is inheritance of basic classes. Furthermore, some of the OOP languages provide multiple inheritance, in particular C++ [Stroustrup, 2013]. However, such approach leads to the same problems, which we have mentioned about frames. That is why we propose another approach to generation of classes of objects. Operations, which were considered in the Example 8, are parts of it. In addition, we will propose another kind of operations, which are modifiers.

**Definition 15.** *Modification function* $m(p_i(A))$ *is a function, which can change somehow a property* $p_i(A)$.

**Definition 16.** *Modifier of object* $A$ *is a vector* $M(A) = (m_1(p_1(A)), \dots, m_n(p_n(A)))$, *where* $m_i(p_i(A))$, $i = \overline{1,n}$ *is a modification function of* $i$-*th property of object* $A$.

**Definition 17.** *Modifier of class of objects* $T$ *is a vector*

$$M(T) = (m_1(p_1(T)), \dots, m_n(p_n(T)), m_1(f_1(T)), \dots, m_k(f_k(T))),$$

*where* $m_i(p_i(A))$, $i = \overline{1,n}$ *is a modification function of* $i$-*th property of class of objects* $T$ *and* $m_j(f_j(T))$, $j = \overline{1,k}$ *is a modification function of* $i$-*th methods of class of objects* $T$

We can divide modifiers on *complete*, *partial*, *generating*, *destroying* and *commutable*, depending on the character of changes of objects or their classes.

**Definition 18.** *Full modifier of object (class) is an object's (class's) modifier, which can simultaneously modify all properties (properties and methods) of particular object (class).*

**Definition 20.** *Partial modifier of object (class) is an object's (class's) modifier, which can simultaneously modify some part of properties (properties and methods) of particular object (class).*

**Definition 21.** *Generating modifier of object (class) is an object's (class's) modifier, which can add some new properties (properties and methods) to specification (specification and signature) of particular object (class).*

**Definition 22.** *Destroying modifier of object (class) is an object's (class's) modifier, which can destroy some properties (properties and methods) of particular object (class).*

**Definition 23.** *Commutable modifier of object (class) is an object's (class's) modifier, which can exchange some properties (properties and methods) of particular object (class) to other properties (properties and methods).*

In addition, there are combined modifiers that simultaneously merge a few different types of modification. We propose a way of creating these combinations (look Table 2).

*Table 2. Combination table of modifiers.*

|   | $F$ | $P$ | $G$ | $D$ | $C$ |
|---|---|---|---|---|---|
| $F$ |  |  |  |  |  |
| $P$ |  |  |  |  |  |
| $G$ |  |  |  |  |  |
| $D$ |  |  |  |  |  |
| $C$ |  |  |  |  |  |

Using Table 2, we can create modifiers that have more complex structure and allow us to describe more difficult transformations of objects and classes of objects.

Generally, modifications of objects and classes of objects give us opportunities to create new classes of objects and thus extend OOP paradigm in this direction.

## Object-Oriented Dynamic Networks

Taking into account advantages and disadvantages of previously considered object-oriented KR models, we tried to propose new object-oriented KRM, which is based on more detail description of objects and classes of objects than frames or scripts, in a sense it is closer to OOP. Let us define it.

**Definition 25.** Object-Oriented Dynamic Network is a 5-tuple $OODN = (O, C, R, E, M)$, where:

- $O$ – a set of objects;

- $C$ – a set of classes of objects, which describe objects from set $O$;

- $R$ – a set of relations, which are defined on set $O$ and $C$;

- $E$ – a set of exploiters, which are defined on set $O$ and $C$;

- $M$ – a set of modifiers, which are defined on set $O$ and $C$.

As you can see, this model uses concepts of object, object's class and operations on them, which were formulated previously. Concerning set of relations, it can contain any relations between objects, classes, including such as *is-a*, *a-kind-of*, *instance-of*, etc., which are common for semantic nets, frames and scripts. Let us consider some examples for more detail explanation of main principals of OODN.

**Example 9.** Suppose we have classes of objects $T(P)$, $T(R)$ and $T(S)$ which describe class of polygons, class of rhombuses, class of squares respectively. Let us define these classes as follows

$$T(P) = (P(P), F(P)) = ((p_1(P), ..., p_3(P)), (f_1(P))),$$
$$T(R) = (P(R), F(R)) = ((p_1(R), ..., p_5(R)), (f_1(R), f_2(R))),$$
$$T(S) = (P(S), F(S)) = ((p_1(S), ..., p_6(S)), (f_1(S), f_2(S))).$$

The meaning of each property and method is defined by the Table 3.

*Table 3. Meaning of properties and methods of classes of objects $T(P)$, $T(R)$, $T(S)$*

| Properties/Methods | Meaning |
|---|---|
| $p_1(P), p_1(R), p_1(S)$ | quantities of sides of figures |
| $p_2(P), p_2(R), p_2(S)$ | sizes of sides of figures |
| $p_3(P), p_3(R), p_3(S)$ | quantities of angles of figures |
| $p_4(P), p_4(R), p_4(S)$ | measure of angles of figures |
| $p_5(R), p_5(S)$ | equality of all sides of figure |
| $p_6(S)$ | equality of all angles of figure |
| $f_1(P), f_1(R), f_1(S)$ | functions of perimeter calculation of figures |
| $f_2(R), f_2(S)$ | functions of area calculation of figures |

In addition, let us consider particular objects of these classes of objects, i.e. rhombus $R_1$ and square $S_1$. The meaning of each property and method is defined by the Table 4 and Table 5 respectively.

*Table 4. Specifications of objects $R_1, S_1$*

| Rhombus $R_1$ | | Square $S_1$ | |
|---|---|---|---|
| Properties | Values | Properties | Values |
| $p_1(R_1)$ | 4 | $p_1(S_1)$ | 4 |
| $p_2(R_1)$ | 2cm, 2cm, 2cm, 2cm | $p_2(S_1)$ | 3cm, 3cm, 3cm, 3cm |
| $p_3(R_1)$ | 4 | $p_3(S_1)$ | 4 |
| $p_4(R_1)$ | 70°,110°,70°,110°, | $p_4(S_1)$ | 90°,90°,90°,90° |
| $p_5(R_1)$ | 1 | $p_5(S_1)$ | 1 |
| × | × | $p_6(S_1)$ | 1 |

*Table 5. Signatures of objects $R_1, S_1$*

| Methods | Values | Methods | Values |
|---|---|---|---|
| $f_1(R_1)$ | $P(R_1) = 4a$ | $f_1(S_1)$ | $P(S_1) = 4a$ |
| $f_2(R_1)$ | $S(R_1) = d_1 d_2 / 2$ | $f_2(S_1)$ | $S(S_1) = a^2$ |

Let us build object-oriented dynamic network for these objects and classes of objects. Clearly, that set of objects is $O = \{R_1, S_1\}$ and set of classes of objects is the following $C = \{T(P), T(R), T(S)\}$. It is obvious, that classes $T(R)$ and $T(S)$ are kinds of class $T(P)$. It is basically know, that the square is a rhombus. According to these facts, we can conclude that sets of relations are the following

$$R = \{R_1 \xrightarrow{\text{instance-of}} T(R), S_1 \xrightarrow{\text{instance-of}} T(S),$$

$$T(R) \xrightarrow{a-kind-of} T(P), T(S) \xrightarrow{a-kind-of} T(P), T(S) \xrightarrow{is-a} T(R)\}.$$

We also can rewrite these relations in the following way

$$R = \{R_1 \in T(R), S_1 \in T(S), T(P) \subseteq T(R), T(P) \subseteq T(S), T(R) \subseteq T(S)\}.$$

Let us define the next set of modifiers $M = \{M_1(T(S)), M_1(T(R)), M_2(T(R)), M_1(T(P)), M_1(R_1)\}$, where: $M_1(T(S))$, $M_1(T(R))$, $M_2(T(R))$, $M_1(T(P))$, transform classes of objects $T(S)$, $T(R)$, $T(R)$, $T(P)$, into classes of objects $T(R)$, $T(L_1)$, $T(S)$, $T(L)$ respectively, $M_1(R_1)$ transforms object $R_1$ to object $L_{1_1}$. Definitions of classes $T(L)$ and $T(L_1)$ will be given later. The parts of OODN for objects $R_1$, $S_1$ and classes of objects $T(P)$, $T(R)$, $T(S)$ are shown on Figure 1 and Figure 2.

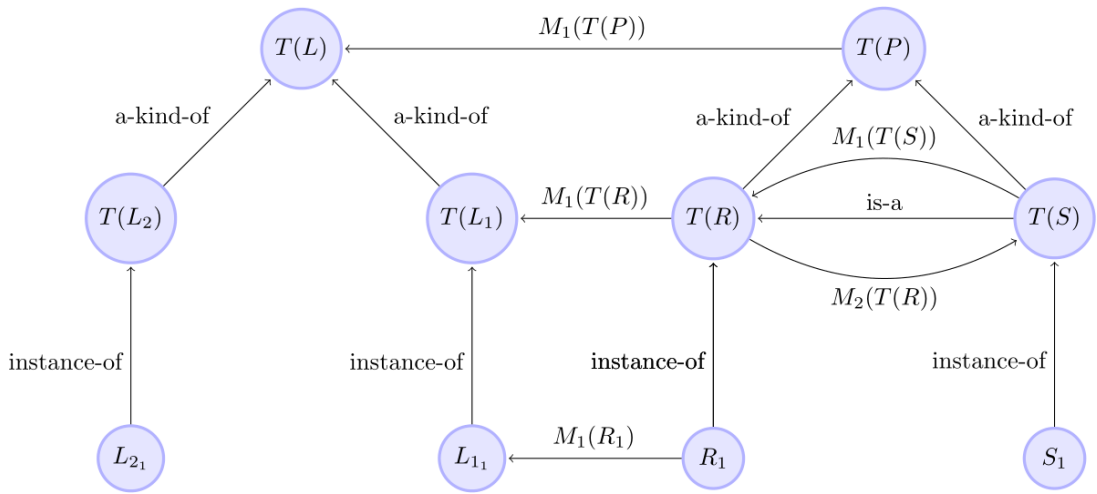

*Fig. 1. Part of OODN of polygons: objects, classes, relations and modifiers.*

On the Figure 1 the part of OODN's graph, which graphically illustrates the structure of the OODN, is drawn. We can divide this graph on left and right parts. Second of them shows the dependencies between classes of polygons and their modifiers, which transform them into other classes. Let us consider modifiers $M_1(T(S))$ and $M_2(T(R))$ in more detail. The first one modifies class of objects $T(S)$ to class of objects $T(R)$, deleting property $p_6(S)$. The second one modifies class of objects $T(R)$ to class of objects $T(S)$, adding property $p_6(S)$. Clearly that they are inverse to each other. These two modifiers illustrate the process of redrawing of geometrical figures and they are examples of partial-deleting modifier.

Now, let us consider modifiers $M_1(T(R))$, $M_1(T(P))$ and $M_1(R_1)$ in more detail. The first one transforms class of objects $T(R)$ to class of objects $T(L_1)$, which is a class of polylines of type $L_1$. This transformation occurs, changing property $p_1(R)$, namely $M_1(T(R)) = m_1(p_1(R)) = m_1(4, sizes) = (3, sizes)$. This modifier is an example of partial modifier. Modifier $M_1(T(P))$ is similar to $M_1(T(R))$. It transforms class of objects $T(P)$ to class of objects $T(L)$ – a class of all polylines. The last one – $M_1(R_1)$ modifies object $R_1$ to object $L_{1_1}$ just as $M_1(T(R))$ modifies $T(R)$ to $T(L_1)$. Analyzing Figure 1, we can conclude that modifiers are kind of transitions between different objects and classes of objects. In such a way, we can model knowledge, which can be modified in time. Furthermore, modifiers form new kind of relations between objects and set of objects.

Generally, these kinds of relations can be represented as *modification-of*. Modifiers can also be considered in temporal context, in particular as future results of modifications.

On the Figure 2 the set of exploiters, which create new objects and classes of objects, using sets $O$ and $C$, without any changing of their elements, is drawn. Analyzing this figure, we can see that it illustrates whole operations on objects and classes of objects from Example 8.

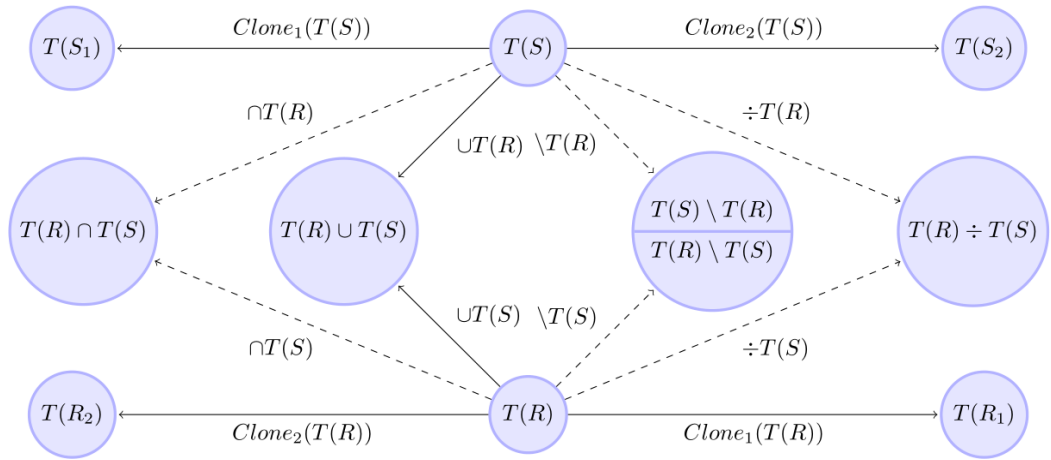

*Fig. 2. Part of OODN of polygons: exploiters.*

There are some edges, which are drown by dotted line. It is because results of corresponding operations do not always exist [Terletskyi, Provotar, 2014]. In contrast to modifiers, which act as transitions, exploiters create new classes of objects based on basic classes from set $C$. In other words, in such a way we can create new classes, which are non-obvious at first glance. Clearly, that there are other types of exploiters, which can be applicable in this example, however there is a question about their generality in comparison with those, which were considered in the Example 8. Analyzing Figure 2, we can conclude that proposed kind of exploiters extend basic set of objects $O$ and classes of objects $C$. In such a way, they increase description's concentration of particular domain.

## Conclusions

This paper contains analysis of such common object-oriented KRM as frames and scripts, which advantages and disadvantages were considered. Furthermore, concepts of object and class were considered from different sides, in particular from OOP's one. The concepts of object and class of objects, which differ from OOP's version, and operations on them, which give us an opportunity to create new sets of objects and new classes of objects, in particular inhomogeneous, were proposed. The operations have set-theoretical nature and are quite general, that gives us a possibility to apply them to any object and class of objects.

The main result of this paper is new object-oriented KRM – Object-Oriented Dynamic Network. It gives us an opportunity to represent knowledge, which can be modified in time, to build new relations between objects and classes of objects and to represent results of their modifications. OODN is based on representation of objects and classes of objects via their properties and methods. It allows us to classify the objects and, in a sense, to build hierarchy of their types. Furthermore, it enables to represent relation of modification between concepts, to build new classes of objects based on existing classes and to create sets and multisets of concepts. Using such model of knowledge representation, we can consider modifications of knowledge and movement through the graph of model as a process of logical reasoning or finding the right solutions or creativity, etc. The proposed approach gives us a possibility to model some aspects of human knowledge system and main mechanisms of human thought, in particular getting a new experience and knowledge. The OODN, in a sense, is similar to OOP languages, but at the same time, it extends classical OOP paradigm, forming new view on the creation of classes of objects. However, despite all advantages, proposed KRM requires further research.

## Authors' Information

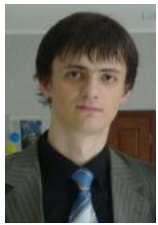
**Dmytro Terletskyi** – 3rd year postgraduate student, Cybernetics Faculty, Taras Shevchenko National University of Kyiv, Kyiv-680, Ukraine; e-mail: dmytro.terletskyi@gmail.com

Major Fields of Scientific Research: Artificial Intelligence, Discrete Mathematics, Programming, Software Engineering.

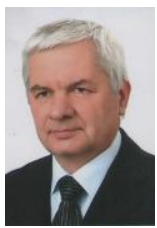
**Alexandr Provotar** – Full Professor, Faculty of Mathematics and Natural Sciences, University of Rzeszów, 35 - 310 Rzeszów, Poland; e-mail: aprowata1@bigmir.net

Major Fields of Scientific Research: Artificial Intelligence, Category theory, Bioinformatics, Nanopharmacology.